\relax
\documentclass[letterpaper]{article} 
\usepackage{ijcai21}  
\usepackage{times}  
\usepackage{helvet} 
\usepackage{courier}  
\usepackage[hyphens]{url}  
\usepackage{graphicx} 
\usepackage{natbib}  
\usepackage{caption} 
\usepackage[switch]{lineno} 
\usepackage{url}
\usepackage{latexsym}
\usepackage{tabularx}
\usepackage{multirow}
\usepackage{diagbox}
\usepackage{graphicx}
\usepackage{extarrows}
\usepackage{algorithm}
\usepackage{algorithmicx}
\usepackage{amsmath}
\usepackage{amsmath,bm}
\usepackage{booktabs}
\usepackage{import}
\usepackage{enumerate}
\usepackage{float}
\usepackage{mathrsfs}
\usepackage{amsfonts}
\usepackage{caption}
\usepackage{xspace}
\usepackage{paralist}
\usepackage{color}
\usepackage{algorithmicx}
\usepackage{algpseudocode}
\usepackage{amsmath}
\usepackage{amsfonts}
\usepackage{arydshln}
\newtheorem{definition}{Definition}

\title{Biomedical Knowledge Graph Refinement with Embedding and Logic Rules}

\author{
Sendong Zhao$^1$
\and
Bing Qin$^1$\and
Ting Liu$^1$\and
Fei Wang$^2$
\affiliations
$^1$Faculty of Computing, Harbin Institute of Technology\\
$^2$Weill Cornell Medical College, Cornell University
\emails
\{sdzhao, bqin, tliu\}@ir.hit.edu.cn,
few2001@med.cornell.edu
}

\begin{document}
\maketitle

\begin{abstract}
Currently, there is a rapidly increasing need for high-quality biomedical knowledge graphs (BioKG) that provide direct and precise biomedical knowledge.
In the context of COVID-19, this issue is even more necessary to be highlighted. However, most BioKG construction inevitably includes numerous conflicts and noises deriving from incorrect knowledge descriptions in literature and defective information extraction techniques. Many studies have demonstrated that reasoning upon the knowledge graph is effective in eliminating such conflicts and noises. This paper proposes a method BioGRER to improve the BioKG's quality, which comprehensively combines the knowledge graph embedding and logic rules that support and negate triplets in the BioKG. In the proposed model, the BioKG refinement problem is formulated as the probability estimation for triplets in the BioKG. We employ the variational EM algorithm to optimize knowledge graph embedding and logic rule inference alternately. In this way, our model could combine efforts from both the knowledge graph embedding and logic rules, leading to better results than using them alone. We evaluate our model over a COVID-19 knowledge graph and obtain competitive results.
\end{abstract}

\section{Introduction}

Collecting structured biomedical knowledge has become a crucial task to provide physicians and doctors with direct and precise biomedical knowledge for decision-making, especially for some new dangerous diseases like COVID-19. 
By the end of July 2020, there had been 646,949 deaths worldwide because of COVID-19, and this number is still increasing \cite{viglione2020medical}. In the face of this global pandemic, great research interest has been attracted, and the literature of COVID-19 is very dynamics and being updated very fast \cite{xiang2020timely}.
According to LitCOVID \cite{chen2020keep}, there have already been more than 47,758 research articles about COVID-19 until Aug. 2020. In reality, it is almost impossible for the readers to keep up with all the articles they are interested in. This makes automatic knowledge graph curation from COVID-9 articles highly demanding. Such knowledge graphs can accelerate the understanding of the transmission and prevention of COVID-19 and help with the battle of COVID-19. 

However, the majority of BioKGs curated with information extraction techniques tend to suffer from low-quality. These BioKGs, which comprise biomedical entities and relations, can be built with NLP techniques including biomedical relation extraction and biomedical named entity recognition. These NLP techniques might bring noises due to their inherent defects. In particular, some non-existing relations might be extracted from biomedical literature. Some biomedical named entities might not be recognized precisely. 
The correct knowledge described in literature can be misextracted by the NLP models, let alone the incorrect descriptions.
There is usually unreliable knowledge in biomedical literature. Unreliable knowledge could be incorrect conclusions, inconsistent and even conflicting answers for the same question in different articles. For example, ``\textit{Young adults have a very low risk of COVID-19}\cite{adams2020medical}" has been proved to be seriously misleading \cite{viglione2020medical}. Besides, there is another example. The evidence in \cite{cardwell2010exposure} indicates that \textit{oral bisphosphonates} are \textbf{not} associated with \textit{esophageal cancer}. However, Green et al. \cite{green2010oral} reached a completely contradictory conclusion that \textit{bisphosphonates} are associated with \textit{esophageal cancer}.

Noises from different ways mentioned above severely diminish the utility of the BioKG. Therefore, it encourages more and more interest in BioKG refinement. However, related studies are all conducted on general knowledge graph like Freebase, Wikidata, and YAGO. These studies can be generally summarized into three  categories that are 1) the web-search based knowledge verification models,  2) fact check models with pattern matching, and 3) hybrid models by combining different resources to verify the knowledge graph. 
Fionda et al. \cite{fionda2018fact} leveraged the knowledge schema to generate evidence patterns and check facts in a knowledge graph with patterns.
Chen et al. \cite{chen2020correcting} proposed a knowledge graph refinement model, which combines lexical matching, graph embedding, soft rule patterns, and semantic consistency checking. 
The above studies have proved that it is effective to verify general knowledge graphs with links of knowledge graph and logic rules.

In a BioKG, each triplet with form (head, relation, tail), denoted as (h, r, t), is composed of a link r and two nodes (h,t). BioKG refinement is to quantize the plausibility of each triplet (h, r, t). 
Knowledge graph embedding has been developed as a promising method for this purpose. This method can learn continuous representations for links and nodes in the knowledge graph to effectively measure the plausibility of triplets with a score function \cite{bordes2013translating}. 
However, one limitation is that they do not consider logic rules which can directly infer knowledge.

In another line of this topic, rule-based approaches \cite{lin2018fact,fionda2018fact,lin2019discovering} are utilized for knowledge graph refinement. Logic rules can be either manually generated by domain experts or mined from the knowledge graph itself. They are used as triplet checking patterns and suggest the correctness of triplets. 
However, logic rules can only cover a small portion of triplets, thus limiting the effectiveness of methods that are purely based on logic rules.

In this paper, we introduce a model for BioKG refinement, which leverages the best of both worlds by combining two clues: 1) knowledge graph embedding which encodes underlying semantics of BioKG; 2) supporting rules and negating-like rules that can be directly applied to support and negate triplets in the BioKG.
In the proposed model, the BioKG refinement is formulated as the plausibility estimation for each triplet. A variational EM algorithm is utilized for training knowledge graph embedding and logic rule inference alternately. Both efforts from the underlying knowledge graph and logic rules can be combined with this alternating process of learning. 
Experimental results over a COVID-19 knowledge graph demonstrate that our model can significantly outperform competitive baselines.

\section{Related Work}

It is critical to verify the correctness of the collected structured knowledge, claim, and statement for real-world usage.
Therefore, this topic has drawn extensive attention from both the research and industrial community.
In general, there are two types of knowledge verification, including fact-checking in free text and fact-checking in the knowledge graph. 
Specifically, fact-checking in free text is to verify textual contents such as claims and statements. Fact-checking in the knowledge graph is to check the correctness of triplets in a given knowledge graph. Here, we survey related work of knowledge verification in the context of knowledge graphs.
There exist three types of approaches for knowledge verification in the knowledge graph.

Knowledge graph embedding models can learn representations of links and nodes in knowledge graph and compute plausibility of triplet candidates with a score function. For the knowledge graph embedding models, we choose five representative models to compare with, including  ConvE \cite{dettmers2017convolutional}, ComplEx \cite{trouillon2016complex}, HolE \cite{nickel2015holographic}, DistMult \cite{yang2014embedding} and  TransE \cite{bordes2013translating}.

Another solution is to apply knowledge graph schema, patterns, or rules to measure the correctness of triplets \cite{lin2018fact,fionda2018fact,lin2019discovering}. 
Lin et al. \cite{lin2018fact} introduced a fact-checking method for knowledge graph with graph fact-checking rules. These rules incorporate expressive subgraph patterns to describe constraints in the knowledge graph. Likewise, Fionda et al. \cite{fionda2018fact} leveraged the knowledge schema to generate schema-level paths as patterns and check facts in a knowledge graph with patterns.
Lin et al. \cite{lin2019discovering} further extended their method in \cite{lin2018fact} for identifying and exploiting similar facts. Therefore, their method is able to check the fact that may not exactly match any known patterns. 

There is a third line of researches which apply multiple resources for fact-checking in knowledge graph \cite{li2017knowledge,chen2020correcting,gad2019tracy,gad2019exfakt,cao2020open}.
Li et al. \cite{li2017knowledge} used various evidence collection techniques to collect evidence from the knowledge graph, the web, and query logs and checked triplets with knowledge fusion. 
A tool was designed to incorporate both evidence in textual sources and the underlying knowledge graph \cite{gad2019tracy}. Similarly, another tool ExFaKT \cite{gad2019exfakt} was proposed to compute explanations over the content of knowledge graphs and textual resources. 
Cao et al. \cite{cao2020open} proposed a probabilistic graphical model to infer the truthfulness of extracted facts from different evidence sources. 


\section{Problem Statement}
A BioKG is a collection of biomedical relational facts, each of which is represented as a triplet (h, r, t).
The problem of noises in BioKG could cause undesired impacts in both research and clinical decision-making process. Therefore, a critical problem of BioKGs is to verify the correctness of triplets.

Formally, given a BioKG $G=(E, R, T, M)$, where $E$ is a set of biomedical entities, $R$ is a set of biomedical relations, $T$ is a set of true triplets (h, r, t) in the BioKG, $M$ is a set of triplets (h, r, t) whose correctness is uncertain in BioKG, 
the goal is to verify the correctness of each triplet candidate $\tau \in M$ 
\begin{definition}
 (Biomeical Knowledge Graph Refinement). Given a BioKG $G$ consists of triplets $(h, r, t)$, \text{biomedical knowledge graph refinement} decides if each triplet in $M$ is true.
\end{definition}
Thus, it is natural to consider this task as a binary classification task that takes as input BioKG $G$ and the triplet candidate $(h, r, t)$ to be verified. 
Following a previous study \cite{nickel2015review}, we can formulate this problem in a probabilistic way. We take the indicator $V_{(h,r,t)}$ of each triplet $(h, r, t)$ as a probabilistic variable. $V_{(h,r,t)} = 1$ represents $(h, r, t)$ is a true triplet, and $V_{(h,r,t)} = 0$ otherwise. Given the BioKG comprise triplets $(h, r, t)$, we aim to estimate the probability of $V_{(h, r, t)}$ for $(h, r, t) \in M$, i.e., $p(V_{(h, r, t)\in M})$. 


\section{Method}
\begin{figure*}[h]
	\centering
	\includegraphics[width=0.8\textwidth]{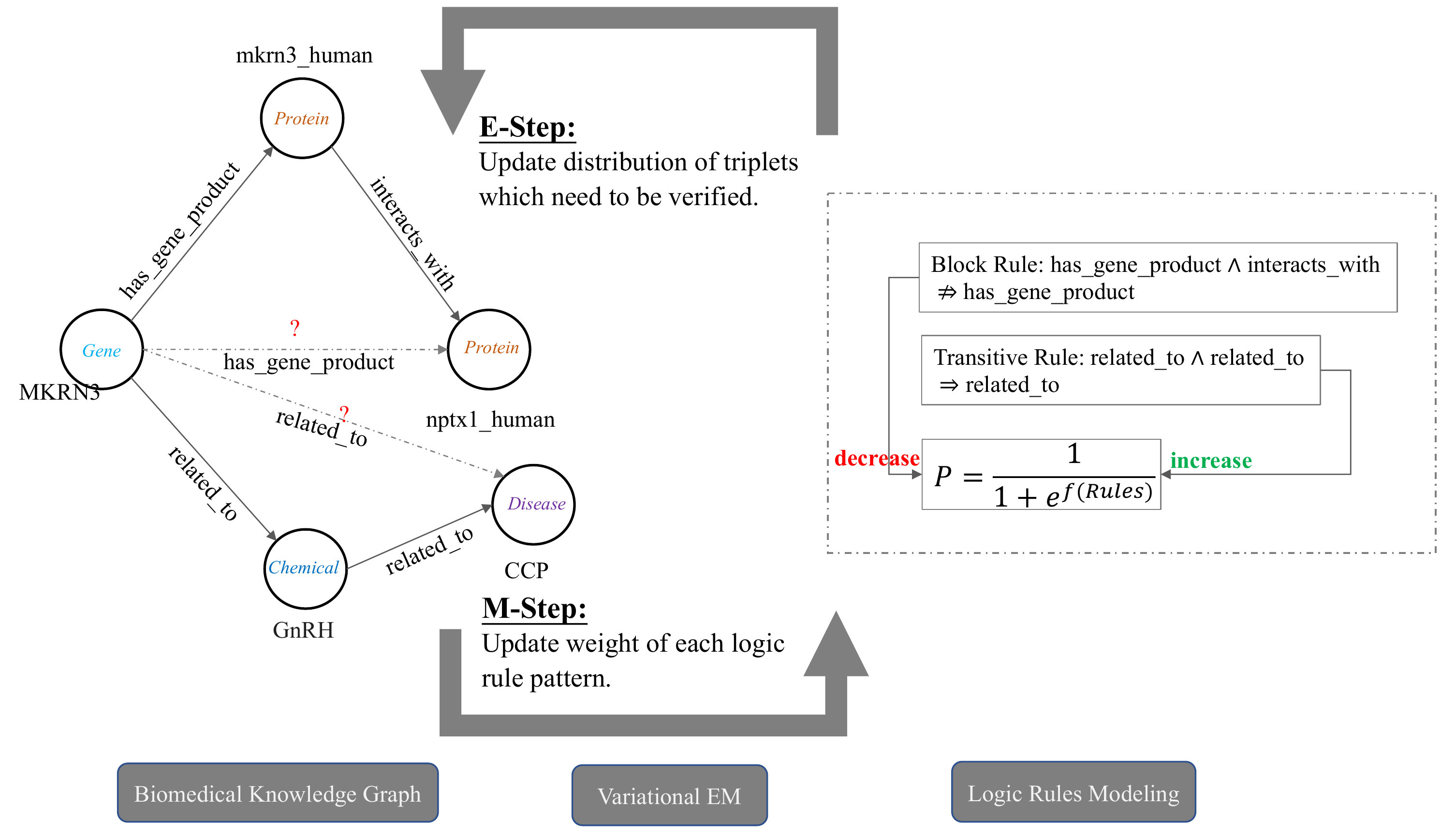}
	\caption{Overview of the BioGRER for combining logic rules and knowledge graph embedding using the variational EM framework.}\label{fig: framework}
	\vspace{-0.2in}
\end{figure*}

In this section, we introduce our proposed approach of \textbf{Bio}medical knowledge \textbf{G}raph \textbf{R}efinement with knowledge graph \textbf{E}mbedding and logic \textbf{R}ules (BioGRER).
This approach incorporates logic rules and the knowledge graph embedding with underlying semantics. 
To better use the logic rules driving from knowledge graph schema, BioGRER considers supporting rules and negating-like rules with a logistic regression model and quantifies the plausibility of given triplets.
The knowledge graph embedding model can learn embeddings of entities and relations with triplets in the knowledge graph. BioGRER incorporates a knowledge graph embedding model to predict the correctness of given triplets with the learned entity and relation embeddings.
These two components are not independent and should augment each other’s capability. Therefore, we jointly train these two components with the variational EM algorithm and allow them alternating between a variational E-step and an M-step. 
In the E step, we use knowledge graph embeddings to predict given triplets to be verified. In the M step, the weights of rules in the logistic regression model are updated based on the true and verified true triplets in BioKG. An overview of our model BioGRER is given in Figure~\ref{fig: framework}.

\subsection{Supporting and Negating-like Rules Modeling}
We apply a logic regression model to incorporate supporting rules and negating-like rules. 
The supporting logic rules enhance the triplet that can be inferred from them. Domain knowledge can be directly encoded in these supporting rules. These supporting logic rules include:
\begin{itemize}
    \item \textbf{Transitive Rules}. A relation $r_k$ is a transitive equivalence of $r_i$ and $r_j$ means that for any three entities $x$, $y$, $z$, if $x$ has relation $r_i$ with $y$, and $y$ has relation $r_j$ with $z$, then $x$ has relation $r_k$ with $z$. We formally define this rule as $\forall x,y,z \in E, V(x,r_i,y)=1 \wedge V(y,r_j,z)=1 \Rightarrow V(x,r_k,z)=1$.
    \item \textbf{Symmetric Rules}. A relation $r$ is symmetric means that for any entity pair $x$ and $y$, if $x$ has relation $r$ with $y$, then $y$ also has relation $r$ with $x$. We formally define this rule as  $\forall x, y \in E, V(x,r,y)=1 \Rightarrow V(y,r,x)=1$.
\end{itemize}

The negating-like rules negate the triplets that are denied by them. In particular, links in the BioKG can be blocked by negating-like rules. These negating-like rules include:
\begin{itemize}
    \item \textbf{Block Rules}. A relation $r_k$ is a block of $r_i$ and $r_j$ means that for any three entities $x$, $y$, $z$, if $x$ has relation $r_i$ with $y$, and $y$ has relation $r_j$ with $z$, then $x$  may not have relation $r_k$ with $z$ with a high rate. We formally define this rule as  $\forall x,y,z \in E, V(x,r_i,y)=1 \wedge V(y,r_j,z)=1 \Rightarrow V(x,r_k,z)=0$.
    \item \textbf{Conflict Rules}. A relation $r_j$ is a conflict of $r_i$ indicates that for any entity pair $x$ and $y$, if $x$ and $y$ have relation $r_i$, then $x$ may not have have relation $r_j$. We formally define this rule as $\forall x,y \in E,V(x,r_i,y)=1 \Rightarrow V(x,r_j,y)=0$.
\end{itemize}

\begin{figure}[h]
	\centering
	\includegraphics[width=0.45\textwidth]{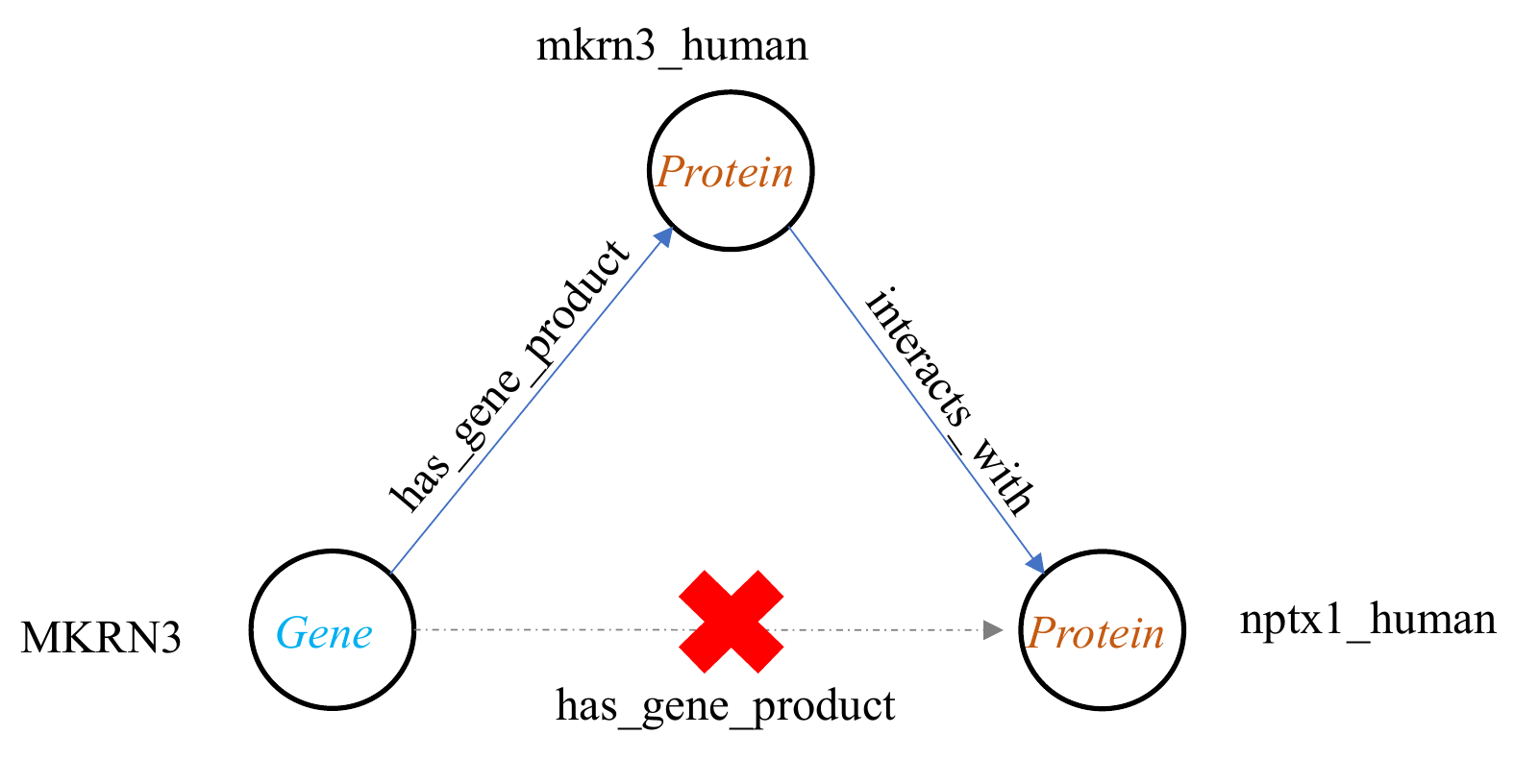}
	\caption{True groundings of block rule for negating triplet candidate (MKRN3, has\_gene\_product, nptx1\_human).}\label{fig: ruleexp}
	\vspace{-0.2in}
\end{figure}

In this study, we use logistic regression to model these above supporting and negating-like rules. The supporting rules should increase the plausibility of triplets. Otherwise, the negating-like rules should decrease the plausibility of triplets. Therefore, we design a positive indicator for supporting rules and a negative indicator for negating-like rules. We have the logistic regression model to quantize the plausibility of each triplet as follows: 
\begin{equation}
    P_{rule}(\tau) = \frac{1}{e^{-f(\tau, T)}+1},
\end{equation}
where $\tau$ is a given triplet which needs to be verified, $T$ is the set of true triplets in BioKG. $f()$ is the linear function to combine logic rules which can be directly applied to the given triplet, which is defined as
\begin{equation}
    f(\tau,T) = \sum_{l\in L}I_lw_lN(l,\tau,T),
\end{equation}
where $l$ is logic rule, $L$ is a set of logic rules $L = \{{l}_{1}^{|L|}$\}, $w_l$ is the weight of the rule $l$, $I_l$ is the indicator for the rule $l$
\begin{equation}
I_l=
\begin{cases}
+1& \text{when l is a supporting rule}\\
-1& \text{when l is a negating-like rule}
\end{cases}
\end{equation}
and $N(l, \tau, T)$ is the number of true groundings of the logic rule $l$ according to the specific rule $l$, the given triplet $\tau$ and neighboring triplets in $T$. 
Figure~\ref{fig: ruleexp} shows an example of true groundings of logic rule for triplet candidate (MKRN3, has\_gene\_product, nptx1\_human). To negate this triplet, the block rule pattern ``$\text{has\_gene\_product} \wedge \text{interacts\_with} \Rightarrow \text{has\_gene\_product is invalid}$" can be applied with the true grounding $V$(MKRN3, has\_gene\_product, mkrn3\_human)=1 $\wedge$ $V$(mkrn3\_human, interacts\_with, nptx1\_human)=1 $\Rightarrow$ $V$(MKRN3, has\_gene\_product, nptx1\_human)=0.

We assume the network consists of all triplets in BioKG as a Markov logic network because a triplet's correctness depends on its neighboring triplets with a certain degree.
Therefore, a triplet is conditionally independent of all other triplets given its Markov blanket in the network.
We define the probabilistic joint distribution of all true triplets $T$ and triplet candidates $M$ as
\begin{equation}
P(V_T, V_M) \geq \prod_{\tau \in T\cup M} P_{rule}(\tau),
\end{equation}
since $P_{rule}(\tau)$ and $P_{rule}(\varsigma)$ are not independent if triplet $\tau$ and triplet $\varsigma$ share the same neighbor(s). In following sections, we will discuss the estimation of this joint distribution.

\subsection{Knowledge Graph Embedding}
Different from the logic rule-based model, the knowledge graph embedding methods learn embeddings of entities and relations with the true triplets $T$ in knowledge graph, and then predict the the correctness of a given triplet $\tau$ with the learned entity and relation embeddings. Formally, head entity $h \in E$, tail entity $t \in E$  and relation $r \in R$ are associated with embeddings $\mathbf{x}_h$, $\mathbf{x}_t$ and $\mathbf{x}_r$. Then the distribution of the given triplet is defined as:
\begin{equation}
    P_{embedding}(\tau)  = \text{Ber}(V(\tau)|f(\mathbf{x}_h, \mathbf{x}_r, \mathbf{x}_t)).
\end{equation}
where Ber stands for the Bernoulli distribution, $f(\mathbf{x}_h, \mathbf{x}_r, \mathbf{x}_t)$ computes the probability that the triplet $\tau = (h, r, t)$ is true, with $f (.,.,.)$ being a scoring function on the entity and relation embeddings. For example in the knowledge graph embedding model TransE \cite{bordes2013translating}, the score function $f$ can be formulated as $\sigma(\gamma -||\mathbf{x}_h + \mathbf{x}_r - \mathbf{x}_t)||)$  according to \cite{sun2019rotate}, where $\sigma$ is the sigmoid function and $\gamma$ is a fixed bias. To learn the entity and relation embeddings, these methods typically treat observed ture triplets and verified triplets as positive examples and the hidden triplets as negative ones. 
In other words, these methods seek to maximize
\begin{equation}
Q(V_T, V_M) = \prod_{\tau \in T \cup M} P_{embedding}(\tau).
\end{equation}
The whole framework can be efficiently optimized with the stochastic gradient descent algorithm.

\subsection{Variational EM}
We introduce the variational EM framework to combine logic rule modeling and knowledge graph embedding together. 
The logistic regression model in Equation (1) models the probabilistic joint distribution of all triplets as in Equation (4).
We can optimize this model by maximizing the log-likelihood of all triplets, no matter true or false in a BioKG. 
However, it is intractable to maximize the objective directly, since it requires integrating over all variables $V_M$ and $V_T$. We instead optimize the variational evidence lower bound (ELBO) of the data log-likelihood, as follows:
\begin{equation*}
\begin{split}
 &\log P(V_T) \geq \log P(V_T) - KL[Q(V_M)||P(V_M|V_T)]\\
 &=\int (Q(V_M)\log P(V_T, V_M)- Q(V_M)\log Q(V_M))dV_M,
\end{split}
\end{equation*}
where KL denotes the KL divergence, and $Q$ represents the variational distribution of triplet candidates to be verified. Equality in the above equation holds when $Q(V_M) = P(V_M |V_T)$. 
We then use the variational EM algorithm \cite{ghahramani2000graphical} to effectively optimize the ELBO. 
The variational EM algorithm consists of an expectation step (E-step) and a maximization step (M-step), which will be called in an alternating fashion to train the model: 1) In the E-step, we infer the posterior distribution of the latent variables, where $P$ is fixed, and $Q$ is optimized to minimize the KL divergence between $Q(V_M)$ and $P(V_M|V_T)$; 2) In the M-step, we learn the weights of the logic rules $w_l$ in logistic region model, where $Q$ is fixed, and $P$ is optimized to maximize the log-likelihood.

\subsubsection{E-step: Inference Procedure}
In the variational E-step, we fix $P$ and update $Q$ by inferring the posterior distribution with mean-field approximation, which is based on the learned embeddings of the knowledge embedding model we have trained, 
\begin{equation}
Q(V_M) = \prod_{\tau\in M} P_{embedding}(\tau).
\end{equation}

Through minimizing the KL divergence between $Q(V_M)$ and the true posterior distribution $P(V_M|V_T)$, the optimal $Q(V_M)$ is computed as
\begin{equation*}
\log Q(V_M) = \mathbb{E}_{Q(V_{MB})}[\log P(V_M|V_{MB})] + const,
\end{equation*}
where $V_{MB}$ is the Markov blanket of $V_M$, which contains the triplets that appear together with $V_M$ in any grounding of the logic rules. If there exists any triplet candidate that is not verified yet in $V_{MB}$, we replace it with a candidate which is predicted as a true triplet by the knowledge graph embedding model. With the equation above, our goal becomes finding a distribution $Q$ that satisfies the condition. To further optimize the objective, we enhance the knowledge embedding model by updating its training dataset with added verified triples, which are predicted by the logistic regression model.
In particular, we first compute $P_{rule}(V_{\tau}|V_{MB})$ for each triplet candidate $\tau$. If $P(V_{\tau}=1|V_{MB}) \geq \delta$ with $\delta$ being a hyperparameter, then we treat $\tau$ as a true triplet and train the knowledge graph embedding model to maximize the log-likelihood $\log P_{embedding}(V_{\tau}= 1)$. Otherwise, the triplet is treated as a negative example. In this way, the knowledge captured by logic rules can be effectively distilled into the knowledge graph embedding model.

\subsubsection{M-step: Learning Procedure}
In the M-step, which is also known as the learning step, we learn the weights of logic rules in the logistic regression model. We fix $Q$ and update the weights of logic rules $w_l$ by maximizing the pseudo-likelihood function, i.e., $\mathbb{E}_{Q(V_M)}[\log P(V_T,V_M)]$.
\begin{equation*}
\begin{split}
&\mathbb{E}_{Q(V_M)}[\log P(V_T,V_M)]\\
& = \mathbb{E}_{Q(V_M)}[\sum_{\tau_M \in M, \tau_T \in T}\log P_{rule}(V_{\tau_M},V_{\tau_T})]\\
&= \mathbb{E}_{Q(V_M)}[\sum_{\tau \in {M\cup T}}\log P_{rule}(V_{\tau}|V_{MB})],
\end{split}
\end{equation*}
where $V_{\tau}$ is the indicator of triplet $\tau$. $V_{\tau} =1$ represents $\tau$ being a true triplet. Otherwise, $V_{\tau} =0$ represents $\tau$ being a false triplet.

In particular, for each true triplet $\tau \in T$, we seek to maximize $P_{rule}(V_{\tau}=1|V_{MB})$. For each triplet candidate $\tau \in M$ which needs to be verified, we treat $P_{embedding}(V_{\tau}=1)$ as target for updating the probability $P_{rule}(V_{\tau}=1|V_{MB})$ to minimize the difference between these two. In this way, the knowledge graph embedding model essentially provides extra supervision to benefit in learning the weights of logic rules.
Besides, triplets candidates distinguished by the knowledge embedding model in E-step are filled into the Markov blanket.
Therefore, by alternating between the variational E-step and an M-step, 
BioGRER allows information sharing between the knowledge graph embedding model and the logistic regression model, thus combines efforts from both sides, as shown in Figure~\ref{fig: framework}.

\section{Experiments}

\subsection{Dataset}
We evaluate the BioGRER on a open available COVID-19 knowledge graph (kg-covid-19)\footnote{https://github.com/Knowledge-Graph-Hub/kg-covid-19/wiki}. This knowledge graph incorporates up-to-date data extracted from biomedical databases and literature, including drug, protein-protein interactions, SARS-CoV-2 gene annotations, concept and publication data from the CORD19 \cite{wang2020cord} data set.
Since the entire graph of this kg-covid-19 is super large, we sample a sub-graph (sub-kg-covid-19) for experiments. 
Table~\ref{tab:data} shows the detailed statistics of the extracted sub-graph of kg-covid-19.
 \begin{table}[h]
\centering
\small
\begin{tabular}{|c|c|c|c|}
\hline
\textbf{Data Set} & \#\textbf{Node} & \#\textbf{Link}& \#\textbf{Relation}\\
\hline
sub-kg-covid-19&3,000&376,505& 107 \\
\hline
\end{tabular}
\vspace{-0.1in}
\caption{The statistics of the sub-kg-covid-19.}
\label{tab:data}
\vspace{-0.2in}
\end{table}

\subsection{Evaluation Metrics and Settings}
\subsubsection{Poisoning Triplet Detection (PTD)} 
We define a task named poisoning triplet detection for BioKG. Specifically, we put new false links to a given knowledge graph and generate new false triplets accordingly. This task is to identify all these new added false triplets. However, annotating these false triplets is tedious for checkers to make sure triplets are genuinely false.
Therefore, how to generate false triplets becomes a big concern. 
To this end, we randomly generate new links and develop new triplets for sub-kg-covid-19. In this process, we check the entire knowledge graph to ensure that these new triplets do not exist in kg-covid-19. 
We take these randomly generated triplets as false triplets and get 10,000 at last. We take 5,000 of these false triplets and 5000 existing true triplets to compose the validation set. We take the rest 5,000 of false triplets and other 5,000 existing true triplets to compose the testing set. We call this testing set as the ``Large" testing set.
Some of new generated false triples may be missing true triplets. However, the number of these cases is relatively small in such a dense graph like kg-covid-19. We sample 100 triplets from 10,000 generated false triplets and manually check them via searching in PubMed and Google. The result is 97\% of them cannot be supported by any evidence from PubMed and Google. In other words,  97\% of them are true false triplets. We take these 97 manually labeled false triplets and other 103 true triplets to compose a testing set of 200 samples. We call it the ``Small" testing set.

\subsubsection{Missing Triplet Prediction (MTP)} 
Missing triplets prediction is a knowledge graph completion task, which is targeted at assessing the plausibility of triples not present in a knowledge graph. For this task, we randomly take out 10,000 triplets from sub-kg-covid-19 as the validation set and take out the other 10,000 triplets as the testing set.

\subsubsection{Metrics}
We compare different methods on the tasks of poisoning triplet detection and missing triplet prediction. We formulate poisoning triplet detection as a triplet classification task. For each triplet in the testing set, our model predicts whether it is true or false.  
Therefore, we follow the standard evaluation metrics for the classification task, i.e., Precision (P), Recall (R) and F-score (F).
We formulate missing triplets prediction as a ranking task.
For each triplet in the testing set, we mask the head or the tail entity, and let each compared method predict the masked entity. Following existing studies \cite{bordes2013translating}, we apply the filtered setting during evaluation. The Mean Rank (MR), Mean Reciprocal Rank (MRR) and Hit@K (H@K) are treated as the evaluation metrics.

\subsubsection{Settings}
We search for all the possible supporting rules and negating-like rules from the observed triplets to generate the candidate logic rules. We compute the empirical precision of each rule for supporting rules, i.e., $p_l=\frac{|R\cap T|}{|R|}$. $R$ is the set of triplets generated with logic rule $l$ in the knowledge graph. They may or may not exist in the knowledge graph. $T$ is the set of true triplets that do exist in knowledge graph. In our study, we assume all triplets in the sub-kg-covid-19 are true. We only select supporting rules whose empirical precision is larger than a threshold $\beta$.
Besides, we compute the empirical precision of each negating-like rule as $p_l=\frac{|D\cap T|}{|D|}$. $D$ is the set of triplets that should be negated with logic rule $l$ in the knowledge graph. However, some of them might exist in the knowledge graph.
We only keep supporting rules whose empirical precision is $1$.
We consider two variants for our approach, where BioGRER uses only $Q$ to infer the plausibility of triplet candidates during the evaluation. In contrast, BioGRER* uses a weighted-sum of $Q$ and $P$, i.e., $Q + \lambda P$. 
We use TransE \cite{bordes2013translating} as the default knowledge graph embedding model to optimize $Q$. We update the weights of logic rules with gradient descent. 
We select the learning rate $\lambda$ for stochastic gradient descent among
\{0.001, 0.01, 0.1\}, the margin $\gamma$ among \{1, 2, 10\}, and the
dimension of embedding $d$ in \{10, 20, 30, 40, 50\} on the validation set. The optimal configurations are
$d = 30$, $\gamma$ = 1, $\beta = 0.3$, $\lambda = 1$.

\subsubsection{Compared Models} 

For BioKG refinement tasks, we compare with the following three types of baselines to evaluate BioGRER.

 \textbf{Rule-based models} apply rules to measure the correctness of triplet candidates. For the rule-based methods, we compare with the Markov logic network (MLN) \cite{richardson2006markov} and the Bayesian logic programming (BLP) method \cite{de2008probabilistic}, which model logic rules with Markov networks and Bayesian networks respectively. Besides, we compare with CHEEP \cite{fionda2018fact} which leveraged the knowledge schema to generate  schema-level  patterns and verify the knowledge graph with patterns.

 \textbf{Knowledge graph embedding models} can learn representations of links and nodes in knowledge graph and compute plausibility of triplet candidates with a score function. For the knowledge graph embedding models, we choose five representative models to compare with, including  ConvE \cite{dettmers2017convolutional}, ComplEx \cite{trouillon2016complex}, HolE \cite{nickel2015holographic}, DistMult \cite{yang2014embedding} and  TransE \cite{bordes2013translating}.

 \textbf{Hybrid models} combine different resources to verify the correctness of triplet candidates. We compare with pLogicNet \cite{qu2019probabilistic}, RUGE \cite{guo2017knowledge} and NNE-AER \cite{ding2018improving}, which are hybrid methods that combine knowledge graph embedding and logic rules. We also compare Node+Path \cite{chen2020correcting}, which combines lexical matching, semantic embedding, soft constraint mining, and semantic consistency checking. 

\subsection{Main Results}
\begin{table*}[h]
\centering
\small
\begin{tabular}{llcccccccccc}
\hline
\multirow{2}{*}{\textbf{Model Type}}&\multirow{2}{*}{\textbf{Model}}&\multicolumn{6}{c}{\textbf{Poisoning  Triplet Detection}}&\multicolumn{3}{c}{\textbf{Missing  Triplet  Prediction}}\\
\cmidrule(lr){3-8} \cmidrule(lr){9-11}
\multirow{2}{*}{}& \multirow{2}{*}{}&\textbf{P@L} & \textbf{R@L}& \textbf{F@L} & \textbf{P@S} & \textbf{R@S}& \textbf{F@S}&\textbf{H@1} & \textbf{MRR} & \textbf{MR}\\
\hline
\multirow{3}{*}{Rule-based} &BLP &1.07 &1.22 &1.14&0.55&1.03 & 0.71&0.03 &0.007 & 4765\\
\multirow{3}{*}{}&MLN                  &1.46 &1.87& 1.63& 1.06&2.06&1.39 &0.03 &0.009 & 4431\\
\multirow{3}{*}{}&CHEEP & 99.73&2.67&5.20&100&3.09& 5.99& 0.03&0.011& 4167\\
\hline
\multirow{5}{*}{KG Embedding}& TransE &6.79 &8.14&7.40&6.06&6.18&6.12 & 3.83 &0.082 & 820\\
\multirow{5}{*}{} &DistMult     &6.43 &7.36 &6.86&6 &5.15& 5.54 & 1.26   & 0.037  & 1125\\
\multirow{5}{*}{} &HolE  & 4.31 & 4.73 &4.51&   4.08&4.12& 4.09   & 0.07   & 0.016  & 2763\\
\multirow{5}{*}{} &ComplEx & 5.07 &5.74&5.38&  4.95  &5.15&   5.05  & 0.12   &0.031   & 1638 \\
\multirow{5}{*}{} &ConvE & 5.11 & 5.33&5.21& 4.9  &5.15&    5.02   & 0.10   &0.027   &1869\\
\hline
\multirow{4}{*}{Hybrid}&RUGE &9.31&8.76& 9.02& 8.43&7.21&7.77 &3.92& 0.082  & 816\\
\multirow{4}{*}{}&NNE-AER & 8.63 &7.87&8.23& 7.14 &6.18& 6.62& 5.64  & 0.103 & 567\\
\multirow{4}{*}{}&pLogicNet &9.77 &8.63 &9.16& 8.86&7.21& 7.95& 4.17 & 0.084 & 817\\
\multirow{4}{*}{}&Node+Path& 16.16&10.21&12.51 & 15.71&11.34 &13.17& 5.37 & 0.096 & 779\\
\hline
\multirow{2}{*}{Our Model}&BioGRER & 67.21 &28.32& 39.85&65.85 &27.83 &39.12& 6.31 & 0.115 & 579\\
\multirow{2}{*}{}&BioGRER* & 75.13 &29.21&42.06& 72.5&29.89& 42.32&6.46&0.117& 564\\
\hline
\end{tabular}
\vspace{-0.1in}
\caption{The overall performance of two tasks over COVID-19 knowledge graph. H@1, P, R and F are \%. (p-value $\leq0.05$)}
\label{tab:overall}
\end{table*}
We conduct experiments for poisoning triplet detection task and missing triplet prediction task with different models. A summary of the results is displayed in table~\ref{tab:overall}. 
The ``P@L", ``R@L" and ``F@L" columns show  the results on the large testing set for the poisoning triplet detection task.  The ``P@S", ``R@S" and ``F@S" columns show the results on the manually labeled small testing set for the poisoning triplet detection task. 
It is apparent that BioGRER significantly outperforms all baseline models, including rule-based models, knowledge graph embedding models, and hybrid models on poisoning triplets detection. 
Compared with rule-based models, BioGRER applies the knowledge graph embedding technique to improve detection performance. In addition to those rules that support triplets, our model encodes all possible rules that negate triplets. This is a different way of encoding logic rules compared with existing rule-based studies. BioGRER outperforms knowledge graph embedding models, as it exploits the knowledge encoded within the logic rules. Moreover, BioGRER consistently performs better than hybrid methods, which shows the superiority of 
negating-like rules for poisoning triplet detection. BioGRER* slightly outperforms BioGRER due to the complementary information captured by $Q$ and $P$. It is reasonable that combining them is better than using single alone. For the missing triplet prediction, BioGRER shares a similar trend with the poisoning triplet detection task but the improvements are limited, indicating that negating-like rules for the missing triplet prediction task are not as effective as for the poisoning triplet detection task. 

\subsection{Analysis of Different Rule Patterns}
 \begin{table*}[h]
\small
\centering
\begin{tabular}{lccccccccc}
\hline
\multirow{2}{*}{\textbf{Logic Rule}}&\multicolumn{6}{c}{\textbf{Poisoning  Triplet  Detection}}&\multicolumn{3}{c}{\textbf{Missing  Triplet  Prediction}}\\
\cmidrule(lr){2-7} \cmidrule(lr){8-10}
\multirow{2}{*}{}&\textbf{P@L} & \textbf{R@L}& \textbf{F@L} & \textbf{P@S} & \textbf{R@S}& \textbf{F@S}&\textbf{H@1} & \textbf{MRR} & \textbf{MR}\\
\hline
Without All &6.79 &8.14&7.40&6.06 &6.18&6.12& 3.83 &0.082 & 820\\
Without Support &71.34 &29.21&41.45 & 69.04 &29.89&13.06&6.13&0.115& 571\\
Without Negate &9.79 &8.62 & 9.17& 9.33&7.21& 8.13& 4.17 & 0.085 & 812\\
\hline
Symmetric  &9.71 &8.62&9.13&8.43&7.21& 7.77& 4.13 & 0.077 & 817\\
Transitive &8.84 &8.62&8.73 &8.23 &7.21&7.69 & 4.15 & 0.081 & 813\\
Conflict  & 17.16&8.93 &11.75 & 15.68& 8.24& 10.80& 4.66 & 0.101 & 764\\
Block   & 70.23&28.89&40.94 &  67.5&27.83&39.41&5.87&0.114& 581\\
\hline
\end{tabular}
\vspace{-0.1in}
\caption{The overall performance of two tasks over COVID-19 knowledge graph. H@1, P, R and F are \%. (p-value $\leq0.05$)}
\label{tab:rule}
\vspace{-0.2in}
\end{table*}
We consider four rule patterns, including two supporting rule patterns and two negating-like rule patterns in our model. To deep understand these rule patterns, we systematically investigate the effect of each rule pattern. The results of varying combinations are presented in table~\ref{tab:rule}. The results show that most rule patterns can lead to improvement compared to the model without logic rules. Moreover, the effects of different rule patterns are quite different. Negating-like rule patterns are more effective than supporting rule patterns. In negating-like rule patterns, the ``Block" rule achieves better performances than the ``Conflict" rule.

\subsection{Effect of Knowledge Graph Embedding Models}
Different knowledge graph embedding models might affect the BioGRER differently. Thus, we compare the performance of BioGRER* with representative knowledge graph embedding models for both tasks. The results are displayed in table~\ref{tab:kge}. ``PTD@L" and ``PTD@S" columns report the F-score of poisoning triplet detection over the large and small testing set. The ``MTP" column presents the H@1 of missing triplet prediction task.   
The results indicate that our model is stable for different knowledge graph embedding models.
\begin{table}[h]
	\centering
	\begin{tabular}{l|c|c|c|}
		\hline
		\textbf{KGE}&\textbf{PTD@L}&\textbf{PTD@S}&\textbf{MTP}\\	
		\hline
		TransE   & 42.06 & 42.32    &6.46 \\
		DistMult&   41.66 &  41.07   &6.46 \\
		ComplEx &  40.91   &  40.31   &6.14 \\
		\hline
	\end{tabular}
	\vspace{-0.1in}
\caption{The performance of the BioGRER* with different knowledge graph embedding models.}\label{tab:kge}
\vspace{-0.2in}
\end{table}

\subsection{Case Study of Logic Rules}
Last but not least, we conduct case studies to better understand the extracted logic rule patterns and specific logic rules. 
 \textbf{Transitive Rule}
The transitive rule ``\textit{tributary\_of}(x,y) $\wedge$ \textit{drains}(y,z) $\Rightarrow$ \textit{part\_of}(x,z)" has an example, i.e., V(\textit{facial vein}, \textit{tributary\_of}, \textit{internal jugular vein})=1 $\wedge$ V(\textit{internal jugular vein}, \textit{drains}, \textit{face})=1 $\Rightarrow$ V(\textit{facial vein}, \textit{part\_of}, \textit{face})=1. Given facial vein being a tributary of internal jugular vein, and internal jugular vein draining from face, we have facial vein being a part of face.
This example shows that the transitive rule is reasonable and benefit the verification of ``part\_of" links in BioKG.
 \textbf{Symmetric Rule}. 
The symmetric rule ``\textit{interacts\_with}(x,y) $\Rightarrow$ \textit{interacts\_with}(y,x)" has an example, i.e.,  V(\textit{CASP1 gene}, \textit{interacts\_with}, \textit{IFITM2 protein})=1 $\Rightarrow$ V(\textit{IFITM2 protein}, \textit{interacts\_with}, \textit{CASP1 gene})=1. Given CASP1 gene interacting with IFITM2 protein, we have IFITM2 protein interacting with CASP1 gene. This rule could increase the plausibility of \textit{interacts\_with}(A,B) links with the existence of \textit{interacts\_with}(B,A).
 \textbf{Block Rule}. 
The block rule ``\textit{has\_gene\_product}(x,y) $\wedge$ \textit{interacts\_with}(y,z) $\Rightarrow$ \textit{has\_gene\_produc}(x,z)=0" has an example, i.e., V(\textit{MKRN3 gene}, \textit{has\_gene\_product}, \textit{MKRN3 human protein})=1 $\wedge$ V(\textit{MKRN3 human protein}, \textit{interacts\_with}, \textit{NPTX1 human protein})=1 $\Rightarrow$ V(\textit{MKRN3 gene}, \textit{has\_gene\_product}, \textit{NPTX1 human protein})=0. 
Given MKRN3 gene, located at the 15th human chromosome, producing MKRN3 human protein and MKRN3 human protein interacting with NPTX1 human protein, it is impossible to have MKRN3 gene producing NPTX1 human protein, which is produced by NPTX1 gene, located at the 17th human chromosome.
This block rule is very effective for negating the triplet (\textit{MKRN3 gene}, \textit{has\_gene\_product}, \textit{NPTX1 human protein}). All knowledge graph embedding baselines predict this triplet as true, while BioGRER predicts it as false because of incorporating this block rule.
 \textbf{Conflict Rule}. 
The conflict rule  ``\textit{has\_primary\_input}(x,y) $\Rightarrow$ \textit{has\_primary\_output}(x,y)=0" has an example, i.e., 
V(\textit{morphine catabolic process}, \textit{has\_primary\_input}, \textit{morphine})=1 $\Rightarrow$ V(\textit{morphine catabolic process}, \textit{has\_primary\_output}, \textit{morphine})=0.
The morphine catabolic process, which is a biological process, takes the morphine as input resulting in the breakdown of morphine. Therefore, morphine catabolic process is not able to have ``\textit{has\_primary\_output}" relation with morphine. This conflict rule is very effective to negate the triplet \textit{has\_primary\_output}(x,y) with the valid \textit{has\_primary\_input}(x,y).

\section{Conclusions}
This paper studied the knowledge graph refinement problem and proposed BioGRER to combine
the advantages of supporting and negating-like logic rules and graph embedding for triplet verification. To jointly model these two clues, we trained them alternatively with the variational EM algorithm. To evaluate the effectiveness of BioGRER, we defined the poisoning triplet detection task and conducted experiments for this task. Besides, we conducted experiments on a standard knowledge graph completion task. 
Experimental results on both tasks demonstrated that our model could significantly outperform competitive baselines. 
Furthermore, BioGRER only utilized the information of the BioKG itself instead of external evidence resources. Therefore, our model can be incremented by adding external evidence resources for better performance.

\bibliography{refer}
\bibliographystyle{named}

\end{document}